%% file: main.tex
\documentclass[sigconf]{acmart} 

\usepackage{amsmath,amssymb}
\usepackage{natbib}
\usepackage{graphicx}
\usepackage{textcomp}
\usepackage{enumitem}
\usepackage{hyperref}
\usepackage{balance} 
\usepackage{color,soul}
\usepackage{xcolor}
\usepackage{comment}
\usepackage{hhline}
\usepackage{caption}
\usepackage{subcaption}
\usepackage{tikz}
\usetikzlibrary{positioning}
\usepackage{pgfplots}
\usepackage{pdflscape}
\usepackage{pgfplotstable}
\pgfplotsset{compat=1.7}
\usepgfplotslibrary{groupplots}
\usepackage{multirow}
\usepackage{algorithm}

\usepackage{comment}
\usepackage{colortbl}

\usepackage[table,xcdraw]{xcolor}

\definecolor{verbalprompt}{RGB}{0,100,200}    
\definecolor{cadence}{RGB}{0,150,50}          
\definecolor{emotion}{RGB}{200,50,50}         
\definecolor{disengagement}{RGB}{128,0,128} 

\makeatletter
\let\@authorsaddresses\@empty
\makeatother

\def\BibTeX{{\rm B\kern-.05em{\sc i\kern-.025em b}\kern-.08em
    T\kern-.1667em\lower.7ex\hbox{E}\kern-.125emX}}

\makeatletter
\def\addlegendimage{\pgfplots@addlegendimage}
\makeatother

\renewcommand\footnotetextcopyrightpermission[1]{} 

\begin{document}

\title[How Children Navigate Successive Robot Communication Failures]{Calling for Backup: How Children Navigate Successive Robot Communication Failures}

\author{Maria Teresa Parreira}
\authornote{The authors contributed equally to this research.}
\email{mb2554@cornell.edu}
\affiliation{
    \institution{Cornell University}
    \state{New York}
    \country{USA}
}
\author{Isabel Neto}
\authornotemark[1]
\email{aineto@fc.ul.pt}
\affiliation{
    \institution{Lasige, Faculdade de Ciências}
    \country{Universidade de Lisboa, Portugal}
}
\author{Filipa Rocha}
\affiliation{
    \institution{Lasige, Faculdade de Ciências, ITI/LarSysInstituto Superior Técnico}
    \country{Universidade de Lisboa, Portugal}
}

\author{Wendy Ju}
\affiliation{
    \institution{Cornell University, Cornell Tech}
    \state{New York}
    \country{USA}
}

\begin{abstract}
How do children respond to repeated robot errors? While prior research has examined adult reactions to successive robot errors, children's responses remain largely unexplored. 
In this study, we explore children's reactions to robot social errors and performance errors. For the latter, this study reproduces the successive robot failure paradigm of Liu et al. with child participants (\textit{N}=59, ages 8-10) to examine how young users respond to repeated robot conversational errors. Participants interacted with a robot that failed to understand their prompts three times in succession, with their behavioral responses video-recorded and analyzed. We found both similarities and differences compared to adult responses from the original study. Like adults, children adjusted their prompts, modified their verbal tone, and exhibited increasingly emotional non-verbal responses throughout successive errors. However, children demonstrated more disengagement behaviors, including temporarily ignoring the robot or actively seeking an adult. Errors did not affect participants' perception of the robot, suggesting more flexible conversational expectations in children. These findings inform the design of more effective and developmentally appropriate human-robot interaction systems for young users.
\end{abstract}


\begin{teaserfigure}
  \centering
  \includegraphics[width=0.85\textwidth]{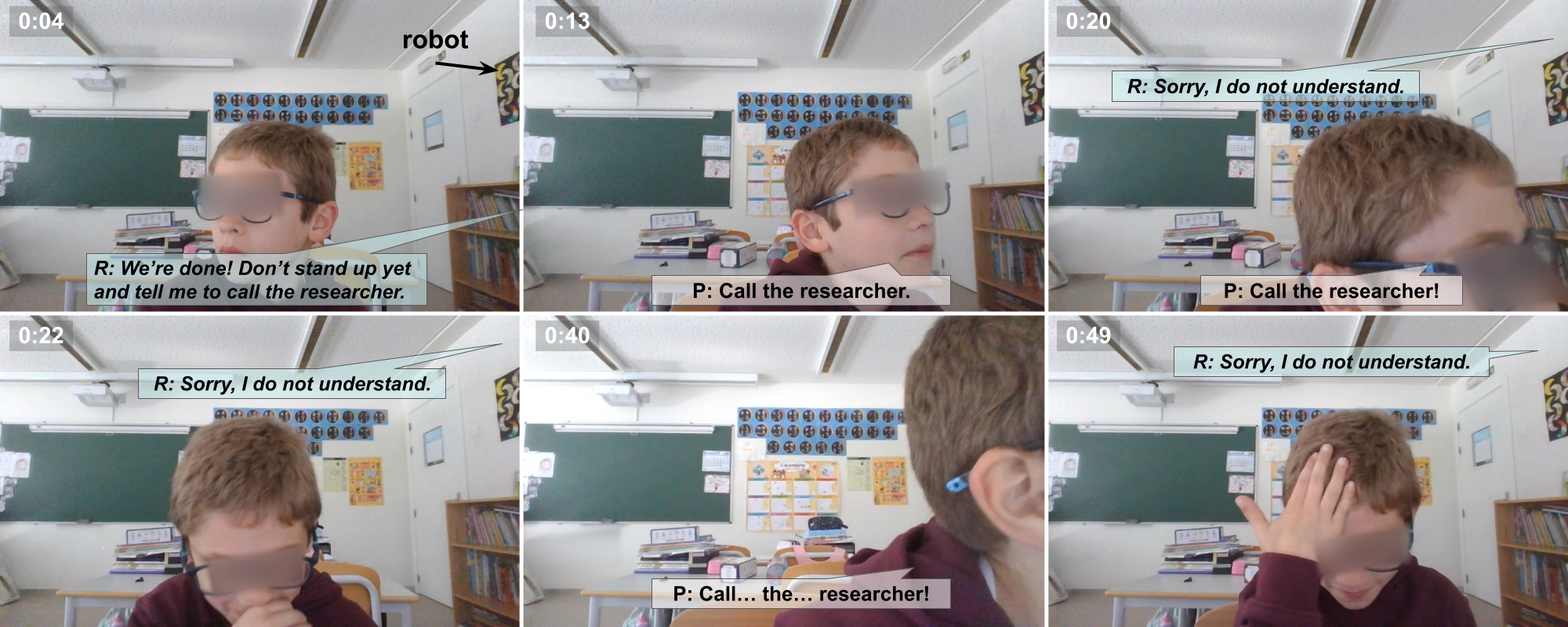}
  \caption{Reactions to successive robot error. After providing instructions to the child, the robot fails to understand their prompt three times. In this vignette, participant reacts to failure by \textcolor{cadence}{moving closer to robot} and using \textcolor{cadence}{verbal adaptations} (demanding tone, slower cadence). After the third error, child exhibits clear signs of \textcolor{emotion}{frustration} and \textcolor{emotion}{humor}, before \textcolor{verbalprompt}{repeating} the command. After this, the researcher enters the room. Supplementary data can be found \hyperref[https://irl.tech.cornell.edu/calling-for-backup/]{in the study repository}.} 
  \label{fig:vignettep32}
\end{teaserfigure}

\maketitle
\pagestyle{plain}

\begin{figure*}
  \centering
  \includegraphics[width=0.95\linewidth]{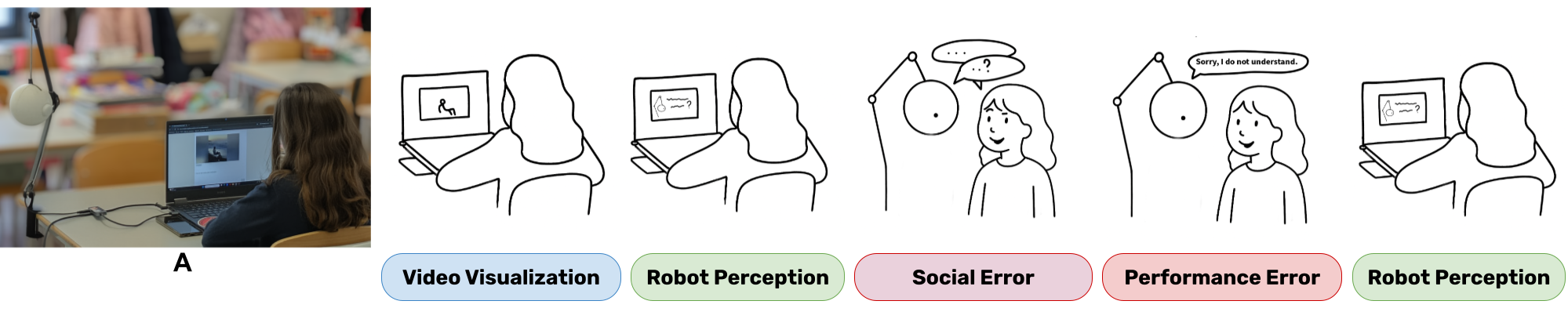}
  \caption{Experimental setup (A) and study protocol. After a short introduction to robot Simon, children watch a series of videos on the laptop. Following this, they fill out a small survey evaluating different dimensions of robot perception. Once they finish, Simon asks them some questions: depending on the condition, it will interrupt their answers (\textit{social error}) or not. After this, Simon fails to understand the participants' request to call the researcher (\textit{performance error}). After 3 successive errors, the researcher is called back into the room. Before ending the session, the child repeats the robot perception survey.}
  \Description{Experimental setup and study protocol.  Child sits in front of laptop, robot is on their left. After a short introduction to robot Simon, children watch a series of videos on the laptop. Following this, they fill out a small survey evaluating different dimensions of robot perception. Once they finish, Simon asks them some questions: depending on the condition, it will interrupt their answers (social error) or not. After this, Simon fails to understand the participants' request to call the researcher (performance error). After 3 successive errors, the researcher is called back in the room. Before ending the session, the child repeats the robot perception survey.}
  \label{fig:schema}
\end{figure*}

\input{sections/01_intro}

\input{sections/02_relatedwork}

\input{sections/03_methodology}

\input{sections/04_results}

\input{sections/05_discussion}

\input{sections/06_conclusion}

\begin{acks}
Funding for this project was granted through Lasige UID/00408/2025 and FCT/PEX  2024.15729.PEX. We thank the schools that collaborated in the data collection. 
\end{acks}

\bibliographystyle{ACM-Reference-Format}
\balance
\bibliography{bibliography.bib,tese.bib}

%

\end{document}

%% file: sections/01_intro.tex
\section{Introduction}
\label{sec:introduction}


The increasing prevalence of robotic systems in diverse domains, from educational companions to healthcare assistants, has expanded their user base to include populations beyond the typical adult demographic~\cite{belpaeme2018review,gordon2016tutor,ros2016motivational, CHEN202teaching}. This includes children, who may have different interaction expectations, communication patterns, and error tolerance compared to adult users~\cite{lemaignan2018pinsoro,Garvey1981turntaking,skantze2017turns,rudenko2024child}.

Despite advances in robotic capabilities, robots remain imperfect and will inevitably make mistakes during interactions. The impact of these errors on users varies depending on user and robot characteristics and contextual factors~\cite{2019sanne,2020kontogiorgosembodiment,kontogiorgos2021systematic}. While prior research in Human-Robot Interaction (HRI) has examined how adults respond to robot failures~\cite{liu2025successive,stiber2023erroraware,bremers2023bad,kontogiorgos2020behavioural,2019sanne}, there remains a critical knowledge gap in understanding how different user populations, particularly children, react to and recover from robot errors. Successive robot failures (when a robot fails multiple times in a row) represent a particularly challenging scenario that is both common in real-world deployments and understudied. Additionally, social errors such as inappropriate interruption during conversation can affect user perceptions of robot likability \cite{aneja2020conversational}. Understanding how children specifically respond to both performance failures and social violations is crucial for designing age-appropriate error detection and recovery strategies.

Recent work by \citet{liu2025successive} explored adult reactions to successive robot conversational failures, revealing complex behavioral patterns including communication strategy adaptations, emotional responses, and eventual interaction abandonment. However, children's cognitive, emotional, and social development differs from adults~\cite{landry2000early,rudenko2024child}, potentially leading to distinct response patterns that require separate investigation.

In this study, we set out to investigate ``\textit{how do children perceive and react to repeated robot error?}''. For this, we reproduce the successive robot failure paradigm with child participants and introduce an additional social error condition to examine how young users respond to both inappropriate robot interruptions and repeated robot conversational failures. Through video analysis and behavioral coding, we characterize children's reactions and compare them to adult responses, providing insights for developing more robust robot error management systems.  Our findings contribute to understanding developmental differences in human-robot interaction and inform the design of child-friendly robotic systems capable of effective error recovery across multiple failure types.

%% file: sections/02_relatedwork.tex
\section{Related Work}
\label{sec:relwork}

\begin{figure*}
    \centering
    \includegraphics[width=0.98\textwidth]{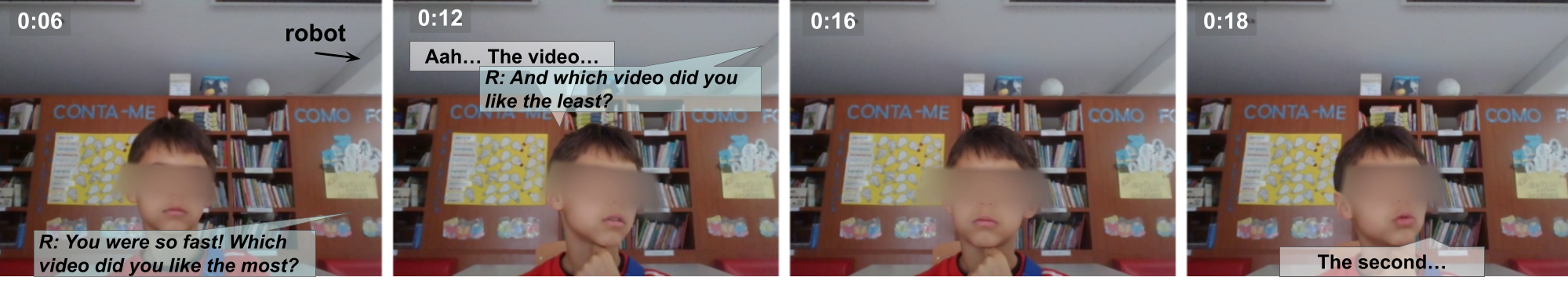}
    \caption{Reactions to social error (interruption). As child starts answering the first question asked by the robot, he gets interrupted. The child does not exhibit external cues of acknowledging this as a socially erroneous behavior. Instead, he thinks for a brief moment before answering the second question. Full social error interaction available as video figure.}
    \Description{Reactions to social error (interruption). As child starts answering the first question asked by the robot, he gets interrupted. The child does not exhibit external cues of acknowledging this as a socially erroneous behavior. Instead, he thinks for a brief moment before answering the second question.}
    \label{fig:vignettep5}
\end{figure*}

We summarize work on developing effective robot error detection and recovery strategies for increasingly diverse user populations.

\vspace{-2pt}
\paragraph{\textbf{Robot Error}}

As opportunities for interactions with robots increase, so do chances for robot error. Robot failures in HRI are a burgeoning field of research \cite{honig2018review,bremers2024usingsocialcuesrecognize,stiber2025roboterrorawarenesshuman}.
The HRI literature typically distinguishes between ``performance errors'' that degrade a user's perception of a robot's intelligence and competence at a task (such as failing to register a command), and ``social errors'' that violate social norms and degrade perceptions of the robot's socio-affective competence (such as inappropriately interrupting a user)~\cite{tian2021taxonomy}.

Research has documented diverse human responses to robot failures across multiple modalities \cite{spitale2024errhri2024challengemultimodal,candon2024react,wachowiak2024errhri}. These responses include verbal adaptations such as reformulating prompts \cite{stiber2025roboterrorawarenesshuman,liu2025successive}, and emotional displays that manifest through facial expressions \cite{kontogiorgos2021systematic,hwang2014reward,2020stiber,2010aronson,bremers2023bad}, changes in gaze patterns \cite{cahya2019socialerror}, and body language adjustments \cite{kontogiorgos2021systematic,2017trung,giuliani2015systematic}. The severity and type of error significantly influence these response patterns \cite{2019sanne}, ultimately impacting user trust and willingness to continue interaction \cite{esterwood2023strikes}. 

Regarding successive errors specifically, \citet{liu2025successive} found that adults adapt communication through reprompting and vocal adjustments, experience emotional progression from confusion to frustration, and may eventually abandon interaction after repeated failures. While similar scenarios with children have been studied, responses appear mixed and complex, warranting further investigation.

\vspace{-2pt}
\paragraph{\textbf{Error in Child-Robot Interaction }}

Children are not simply miniature adults; their ongoing neurophysical and cognitive development creates distinct conditions for robot interaction~\cite{rudenko2024child}.  Research indicates children tend to anthropomorphize robots more than adults~\cite{balpaeme2013cri} and may demonstrate greater tolerance of robot limitations. This tolerance manifests in various ways: \citet{lemaignan2015wrong} found children were more interested in misbehaving robots than predictable ones, while \citet{yadollahi2018deitic} suggest robot mistakes can enhance children's participation by allowing them to correct inaccuracies.

Children demonstrate remarkable perseverance when communicating with robots, finding ways to explain and excuse robot failures~\cite{Turkle2006artifacts}. Additionally, children may be more oblivious to certain types of errors~\cite{Clark_2009}, and technical limitations that are obvious to engineers may go undetected when children believe robots possess human-like capabilities~\cite{nalin2011children}. However, the picture is not uniformly positive: some studies show that repeated robot failures can increase frustration~\cite{tanaka2012caregiving} or decrease trust~\cite{geiskkovitch2019trust}. This mixed evidence highlights the need for systematic investigation of how children respond to successive robot failures across different error types.

%% file: sections/03_methodology.tex
\section{User Study}\label{sec:methods}

To study the effect of conversational failures in children, we followed an adjusted protocol from \citet{liu2025successive}. Our study investigates two dimensions of failures: interruption (social error) and failure to understand (performance error). For social error, we tested two conversational conditions: in the \textit{Interruption} condition, the robot (wizarded) did not wait for the participant to finish answering its previous question before moving forward. In the \textit{Control} condition, the robot did not interrupt the child. Additionally, we studied children's behavior patterns during successive robot performance errors (\textit{failure to understand}).

\subsection{Study Protocol}
\label{subsec:protocol}

The study setup and stages are shown in \autoref{fig:schema}. 
Participants were recruited through the local school network. All children had a written consent form signed by their legal guardian and were informed they could quit the activity whenever they wanted. Initially, participants entered an isolated classroom where they were introduced to ``robot Simon'' (Nodbot). They were asked to sit on a chair, facing a laptop, and follow the instructions on a survey and provided by the robot. After a short chat with the robot (``\textit{Hello, what is your name?''}, ``\textit{Have you talked to a robot before?}''), participants filled out demographic information and were then asked to watch a balanced set of 6 short videos which included robots and humans, with successful or unsuccessful (e.g. failure) outcomes (full list in Supplementary Material). After this, participants filled out a small questionnaire on perceptions of the robot (more details in \autoref{subsec:measures}). Participants would then let Simon know (by speaking aloud) that they had finished. The robot would ask them a series of questions (``\textit{Which video did you like the most?}'', ``\textit{And the least?}''), during which the robot exhibited \textit{Interruption} or \textit{Control} behavior. After this, the robot would instruct participants to request it (the robot) to call the researcher back into the room (``\textit{We are done! Do not stand up yet and tell me to call the researcher.}'')
The robot, which was wizarded, was programmed to cause a failed interaction. The interaction between the participant and Simon was expected to proceed as follows:
\begin{enumerate}
    \item Participant asks Simon to call the researcher.
    \item Simon replies \textit{``Sorry I do not understand''} (Error I)
    \item Participant alerts Simon again.
    \item Simon replies \textit{``Sorry I do not understand''} (Error II)
    \item Participant alerts Simon again.
    \item Simon replies \textit{``Sorry I do not understand''} (Error III)
    \item Participant alerts Simon again.
    \item Simon replies \textit{``Okay I will call the researcher''}.
\end{enumerate}

Finally, the researcher entered the room. Children were asked to fill out the questionnaire a second time to evaluate the effect of the interaction on robot perception (researcher was not in the room). Finally, participants were debriefed, and the deception was explained. No compensation was provided. The full procedure took around 7 minutes to complete. This data was collected under Cornell University IRB exempt protocol \#IRB0010006. The study procedure was approved by the Ethical Committee of Universidade de Lisboa.

\begin{figure*}
    \centering
    \includegraphics[width=0.9\textwidth]{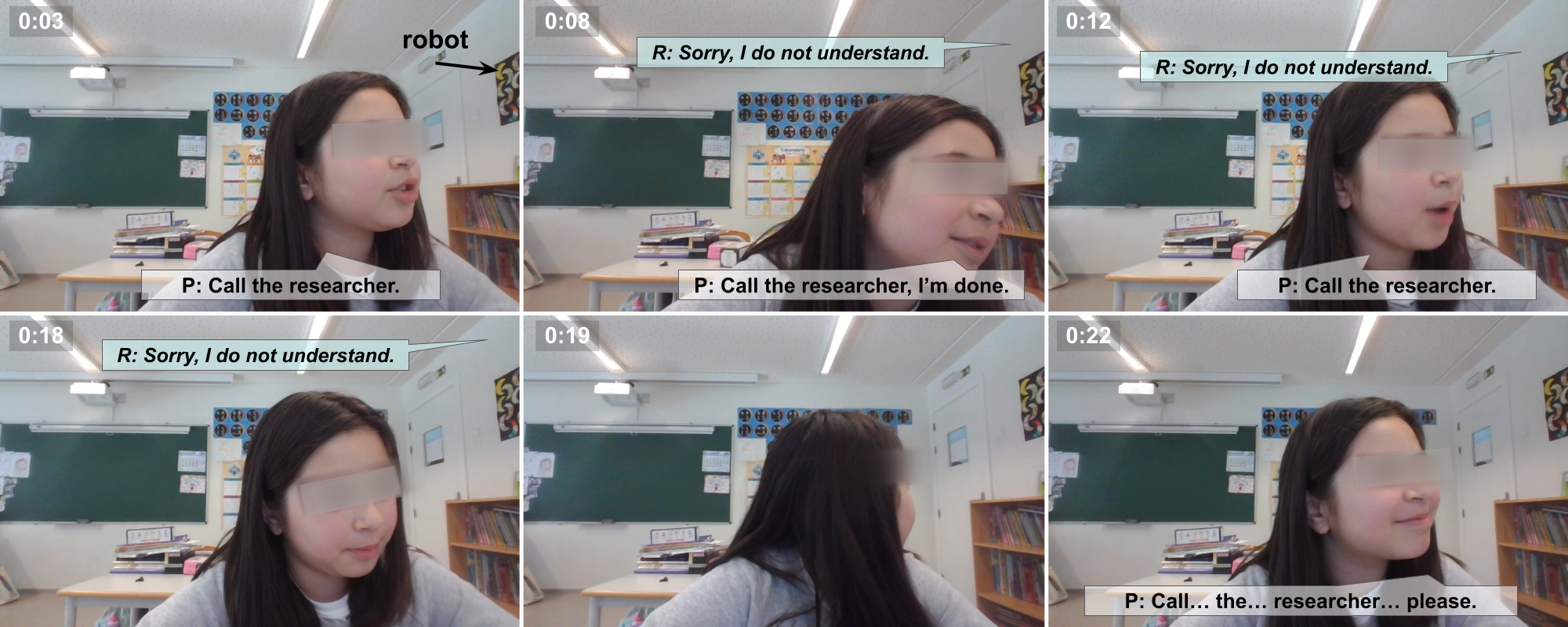}
    \caption{Responses to successive performance error. Child responds to robot error by repeating the command in a \textcolor{verbalprompt}{longer prompt}, \textcolor{cadence}{moving closer to the robot}, with displays of \textcolor{emotion}{frustration} after the third error, as well as \textcolor{disengagement}{looking for the researcher}. After a longer response time, the child finally repeats the command, resorting to \textcolor{verbalprompt}{politeness} and \textcolor{emotion}{smiling} humorously.}
    \label{fig:vignettep30}
\end{figure*}

\subsection{System implementation}
\label{subsec:system}

``Simon'' is a Nodbot \footnote{Based on the design by Rei Lee: \url{https://infosci.cornell.edu/~reilee/NodBot/}}. The robot can be controlled over 2 axes (longitudinal and transverse) through servomotors and can emit sound through a Bluetooth speaker. The interactions in the room were recorded on 2 cameras, a room view and the laptop webcam. The interaction was wizarded. Participants answered the surveys and watched videos on a separate laptop.

\subsection{Measures}
\label{subsec:measures}

To assess children's perceptions of the robot, we administered a small questionnaire consisting of five questions measuring key dimensions of human-robot interaction. All questions used a 5-point Likert scale (1 = not at all, 5 = very much) and were presented in Portuguese. Table~\ref{tab:questionnaire} presents the complete set of questions and their corresponding measured dimensions. These were based on previous literature on measuring robot perceptions \cite{kidd2004perception,neto2024touching,bartneck2008perception} and target dimensions of willingness to continue interacting, competence, trust, social acceptance, and likability. The Cronbach's alpha for these questions is 0.8.

\begin{table}[htbp]
\centering
\small
\caption{Robot perception questionnaire items and measured dimensions.}
\label{tab:questionnaire}
\begin{tabular}{p{0.57\linewidth}|p{0.3\linewidth}}
\hline
\textbf{Question} & \textbf{Dimension} \\
\hline
From 1 to 5, how much would you like to \textit{talk again} with robot Simon? & \textbf{Willingness to continue interacting} \\
From 1 to 5, how much do you think robot Simon \textit{ knows how to talk} with you? & \textbf{Competence} \\
From 1 to 5, how much do you \textit{trust} robot Simon? & \textbf{Trust} \\
From 1 to 5, how much would you like to \textit{be friends} with robot Simon? & \textbf{Social acceptance} \\
From 1 to 5, how \textit{pleasant} do you find robot Simon? & \textbf{Likeability} \\
\hline
\end{tabular}
\end{table}

We calculated difference scores (post-interaction minus pre-interaction) for each participant and compared these changes between the two experimental conditions using independent samples tests. Based on pilot testing, we administered the post-interaction questionnaire after the successive failure sequence rather than immediately following the interruption behavior, as the brief interaction duration (three questions) would have fragmented the experimental flow too frequently, and preliminary results showed minimal differences between these measurement points. This approach captures the compounded effect of both interruption behavior and successive performance errors on robot perception.

\subsection{Reaction Annotation}
\label{subsec:annotation}

The videos from each participant (laptop camera) were analyzed by researchers to identify common response patterns. Annotations were created using ELAN \footnote{\url{hhttps://archive.mpi.nl/tla/elan}}. 
For \textbf{social error }(interruption), we annotated instances wherein a) child \textit{continues responding} to first question (asked before interruption); b) child answers \textit{new question}; c) child \textit{disengages} from interaction.

For \textbf{performance error} (successive failure to understand), we adapted codes from \cite{liu2025successive}, which are documented in a codebook, available as Supplementary Material. The codes include \textbf{\textcolor{verbalprompt}{verbal reprompting}} (e.g., using more specific words in prompts, using simpler prompts), \textbf{\textcolor{cadence}{modifying tone or cadence}} (e.g. speaking slower, using a demanding tone), \textbf{\textcolor{emotion}{emotional displays}} (e.g. confusion, frustration) and \textbf{\textcolor{disengagement}{disengagement}} from the interaction (e.g., stand up, look for researcher).

\begin{figure}
    \centering
    \includegraphics[width=0.95\linewidth]{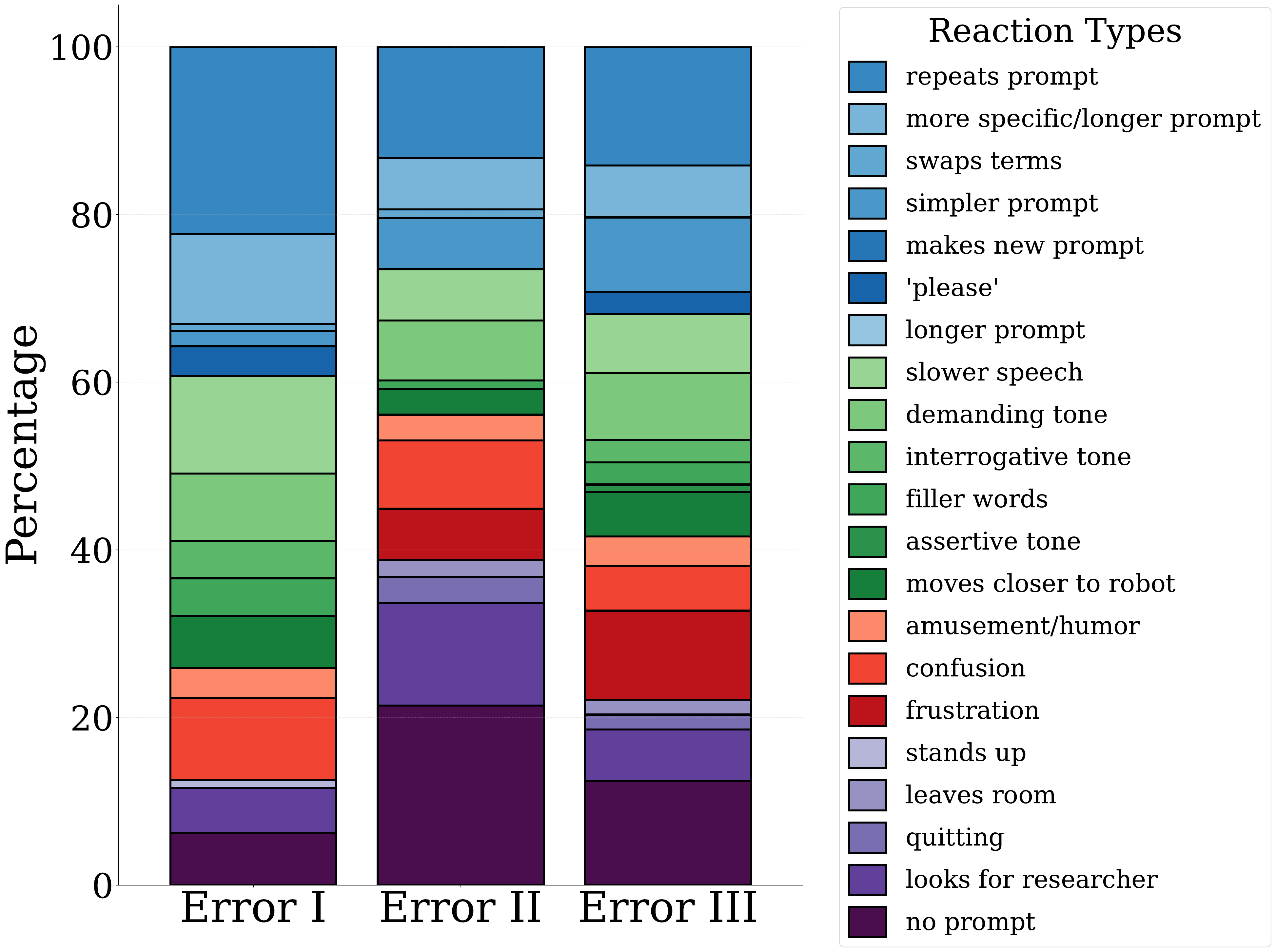}
    \caption{Distribution of children's behavioral responses across successive robot errors. Shows the relative frequency of different reaction types.}
    \Description{Distribution of children's behavioral responses across successive robot errors. Shows the relative frequency of different reaction types. Error I: mostly reprompting strategies and verbal adaptation. Error II, a lot of interaction abandonment. Error III: more evenly distributed across the 4 error categories.}
    
    \label{fig:reactions_by_error}
\end{figure}



\subsection{Participants}

The study included 59 participants across two experimental conditions for social error: 30 in the \textit{Interruption} condition and 29 in the \textit{Control} condition. All participants experienced successive performance error behavior. Participants were aged 8-10 years,(8 yo.: n=35, 9 yo.: n=21, 10 yo.: n=3), 32 boys and 27 girls. Nearly all participants were Portuguese nationals, with three exceptions: one Ukrainian, one American, and one Brazilian participant.
Due to technical issues or background noise, we excluded 6 participants from the social error analysis and 9 participants from the performance error reaction analysis.

%% file: sections/04_results.tex
\section{Results} \label{sec:results}

Below, we show results on how social and performance errors affect perception of the robot and carry out an extensive and comparative behavior analysis to successive robot performance errors. We provide a supplementary video as additional material.

\begin{figure}
    \centering
    \includegraphics[width=0.9\linewidth]{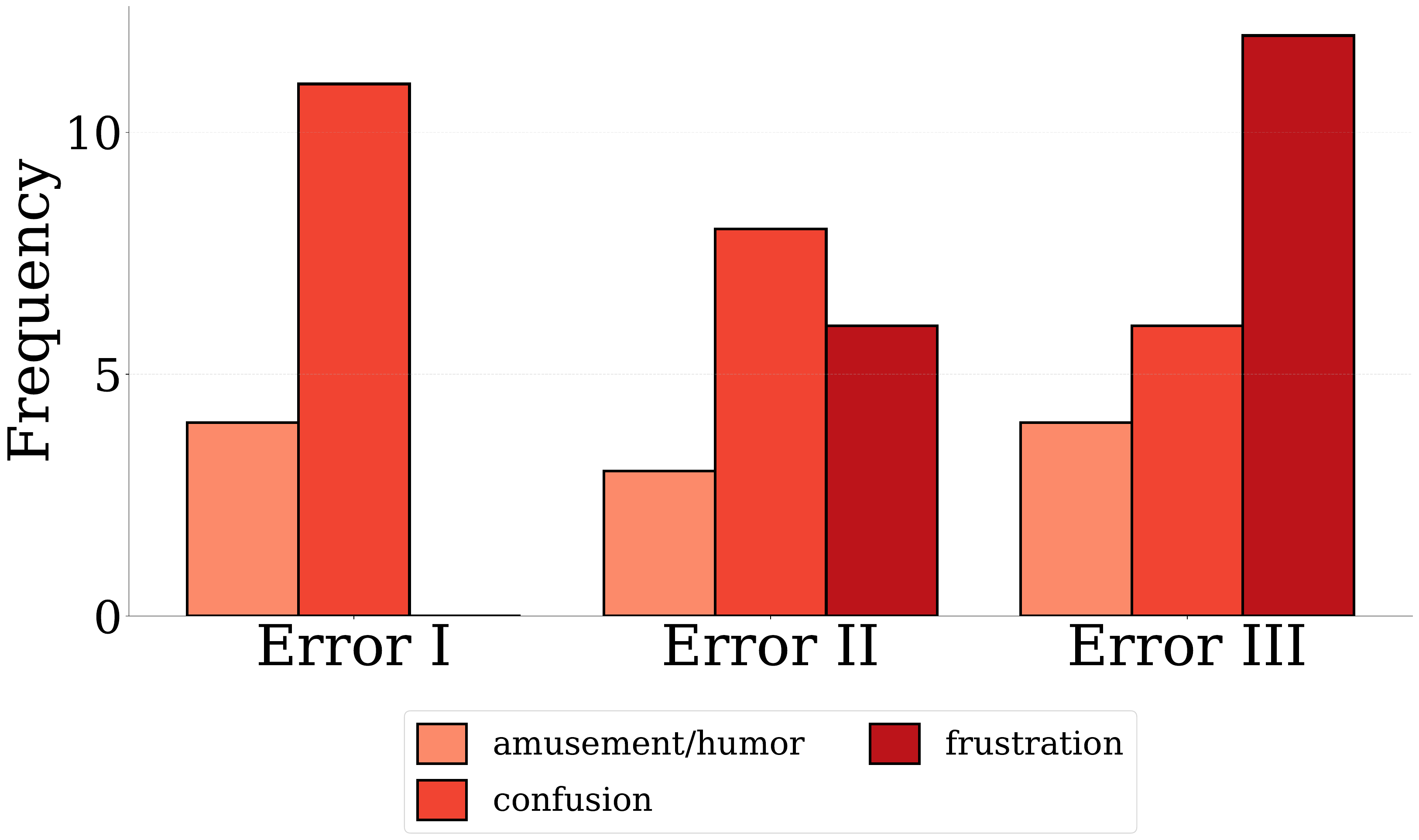}
    \caption{Frequency of emotional reactions across successive robot errors. Confusion peaks after Error II, while frustration progressively increases and becomes dominant by Error III. Amusement/humor remains relatively consistent across all errors.}
    \Description{Frequency of emotional reactions across successive robot errors. Confusion peaks after Error II, while frustration progressively increases and becomes dominant by Error III. Amusement/humor remains relatively consistent across all errors.}
    \label{fig:emotion_frequency}
    \vspace{-10pt}
\end{figure}

\subsection{Impact of Robot Error on Perception}
\label{subsec:results_perception}

For the social error interaction, in the \textit{Control} condition, 3 children did not engage with the robot, and the remaining 25 participants answered the robot questions naturally. In the \textit{Interruption} condition, 21 children ($84\%$) shifted their answer to the new question asked, not acknowledging the interruption (see \autoref{fig:vignettep5} for example), and only 2 children continued responding to the previous questions. Two children did not engage with the robot.

We tested whether interruption affected robot perceptions by evaluating score differences for each condition.
We assessed the normality of the difference score distributions using Shapiro-Wilk tests. As the assumption of normality was violated for all questionnaire items (all $p < 0.05$), we employed non-parametric Mann-Whitney U tests to compare difference scores between the \textit{Interruption} and \textit{Control} conditions. After applying Bonferroni correction for multiple comparisons ($\alpha = 0.01$), there were \textit{no statistical differences in robot perception} between conditions (full results in Supplementary Materials). Given the absence of significant differences between experimental conditions, subsequent analyses are conducted on the combined dataset, treating all participants as a single group regardless of their assigned condition.

To examine whether the robot interaction itself influenced children's perceptions regardless of experimental condition, we conducted paired comparisons between pre- and post-interaction responses across all participants ($n = 52$). Shapiro-Wilk tests indicated non-normal distributions for all difference scores (all $p < 0.05$), therefore Wilcoxon signed-rank tests were employed for each questionnaire item. None of the comparisons approached statistical significance after applying Bonferroni correction for multiple comparisons ($\alpha = 0.01$), indicating that the erroneous robot interaction had no meaningful impact on children's perceptions of the robot across any measured dimension.

\subsection{Reactions to Successive (Performance) Error}
\label{subsec:results_behavior}

We analyzed the videos and noted behavioral patterns in children's responses to successive performance errors (failure to understand), which we explain below and illustrate through vignettes and a supplementary video figure. We also compare reaction patterns with those reported in \citet{liu2025successive}, to infer how these generalize.

Successive error interactions with the robot averaged to be $36.28\pm16.07 (M\pm SD)$ seconds. A total of 402 annotations were used for analysis (more details in Supplementary Materials).

\subsubsection{Children's Behavioral Responses} \label{sec:reactions}

We identified distinct patterns in children's responses to successive robot failures across verbal, emotional, and engagement dimensions. \autoref{fig:reactions_by_error} shows the distribution of reaction types across successive errors.

\paragraph{\textbf{Verbal Communication Strategies}}
Children employed various reprompting strategies, adjusting both content and delivery. They modified prompts (repeating, simplifying, or adding specificity), altered their vocal tone (slower speech, demanding or interrogative tones), and in some cases shifted from assertive commands (``Call the researcher.'') to polite requests (``Could you call the researcher, please?'') (\autoref{fig:vignettep32},\autoref{fig:vignettep30}). 


\paragraph{\textbf{Emotional Response Evolution}}
Emotional displays evolved significantly across errors. Confusion dominated after Error I, remained present in Error II, while frustration peaked by Error III (Figure~\ref{fig:emotion_frequency}). Amusement appeared consistently but remained relatively low across all errors.

\begin{figure}
    \centering
    \includegraphics[width=0.9\linewidth]{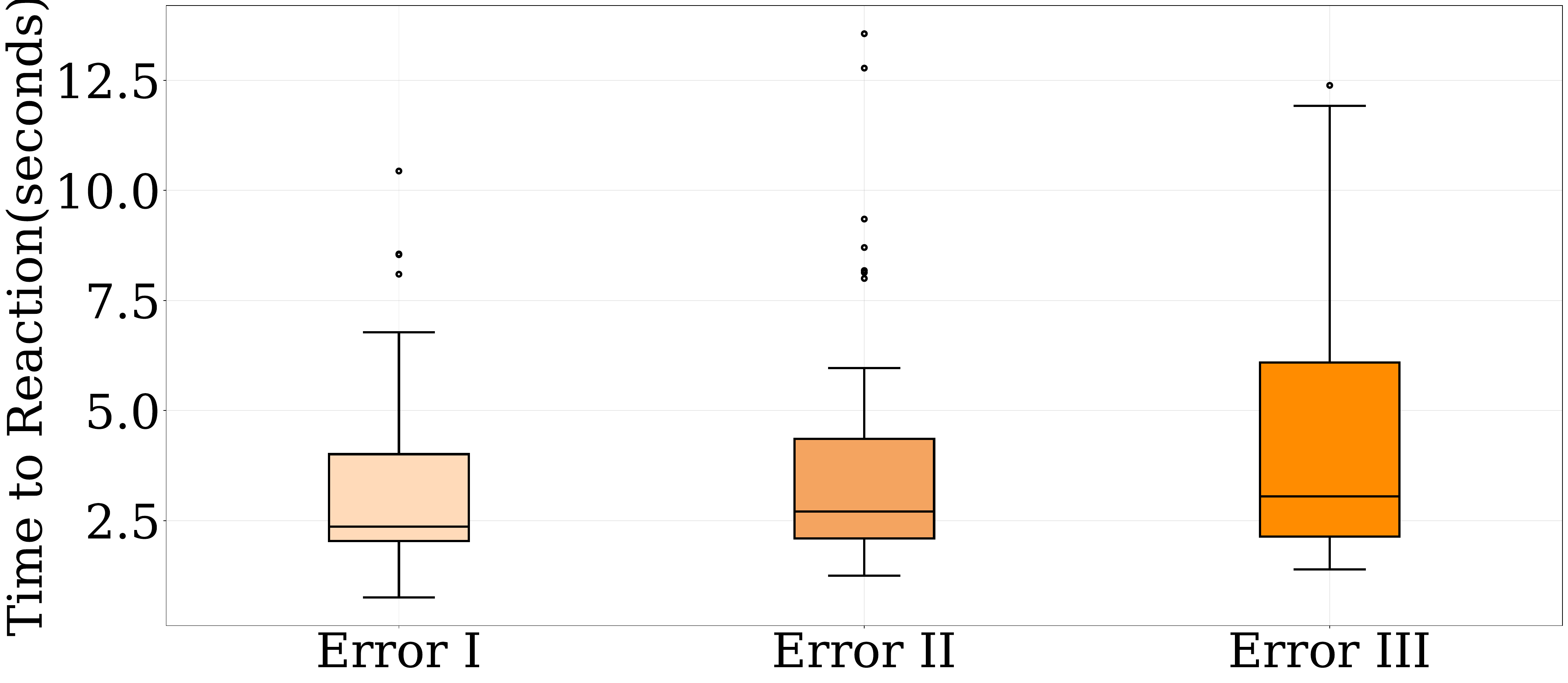}
    \caption{Response latencies across successive robot errors. Response times increase and become more variable with successive errors, with Error III showing the longest delays and greatest variability.}
    \Description{Response latencies across successive robot errors. Response times increase and become more variable with successive errors, with Error III showing the longest delays and greatest variability.}
    \label{fig:response_times}
\end{figure}

\begin{figure}
    \centering
    \includegraphics[width=0.9\linewidth]{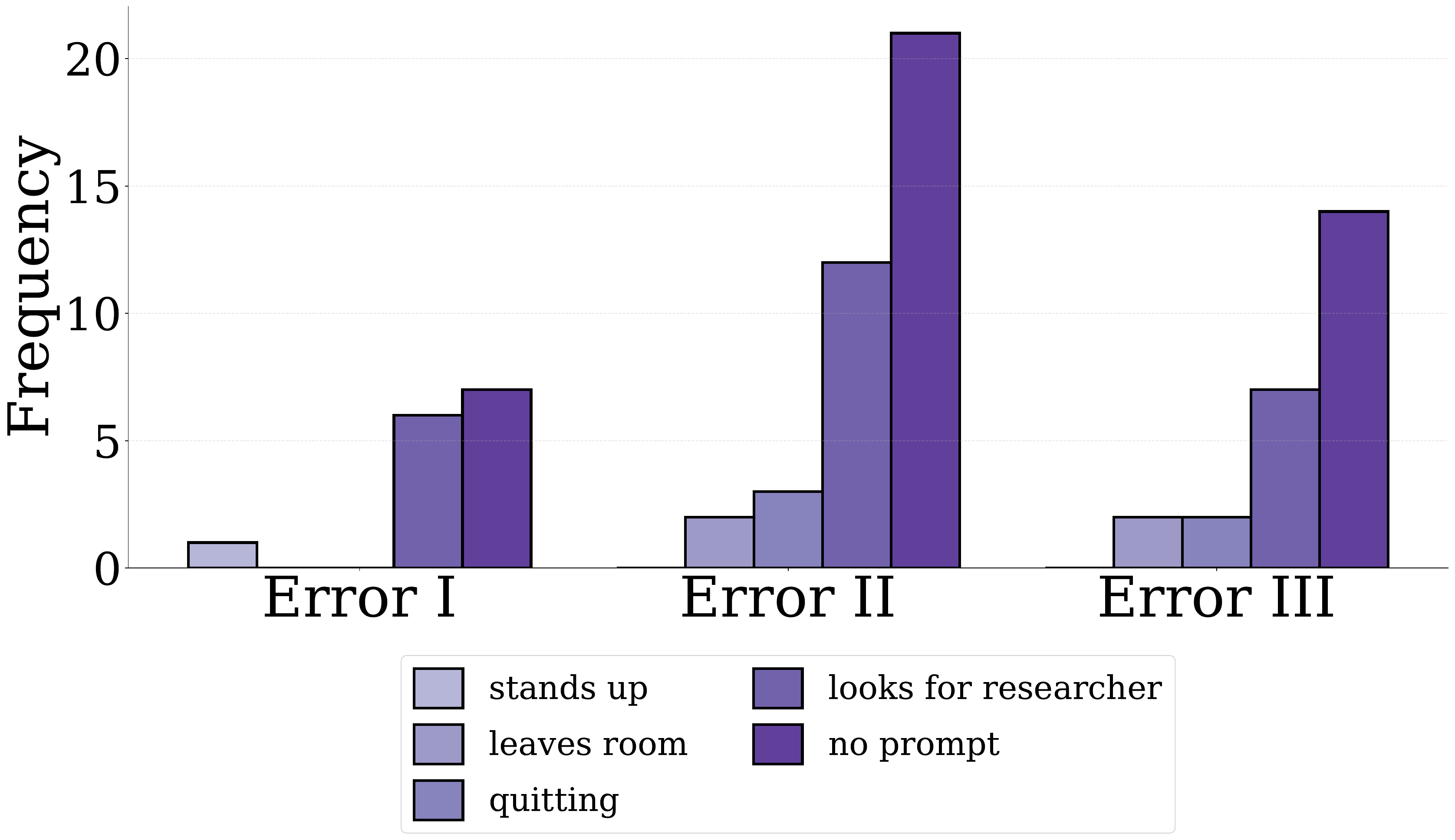}
    \caption{Frequency of disengagement behaviors across successive robot errors. Shows progressive increase in disengagement strategies, with ``no prompt'' and ``looks for researcher'' being most prevalent by Error III. Other behaviors include standing up, leaving room, and complete interaction abandonment.}
    \Description{Frequency of disengagement behaviors across successive robot errors. Shows progressive increase in disengagement strategies, with ``no prompt'' and ``looks for researcher'' being most prevalent by Error III. Other behaviors include standing up, leaving room, and complete interaction abandonment.}
    \label{fig:disengagement_frequency}
\end{figure}

\begin{figure*}
    \centering
    \includegraphics[width=0.9\textwidth]{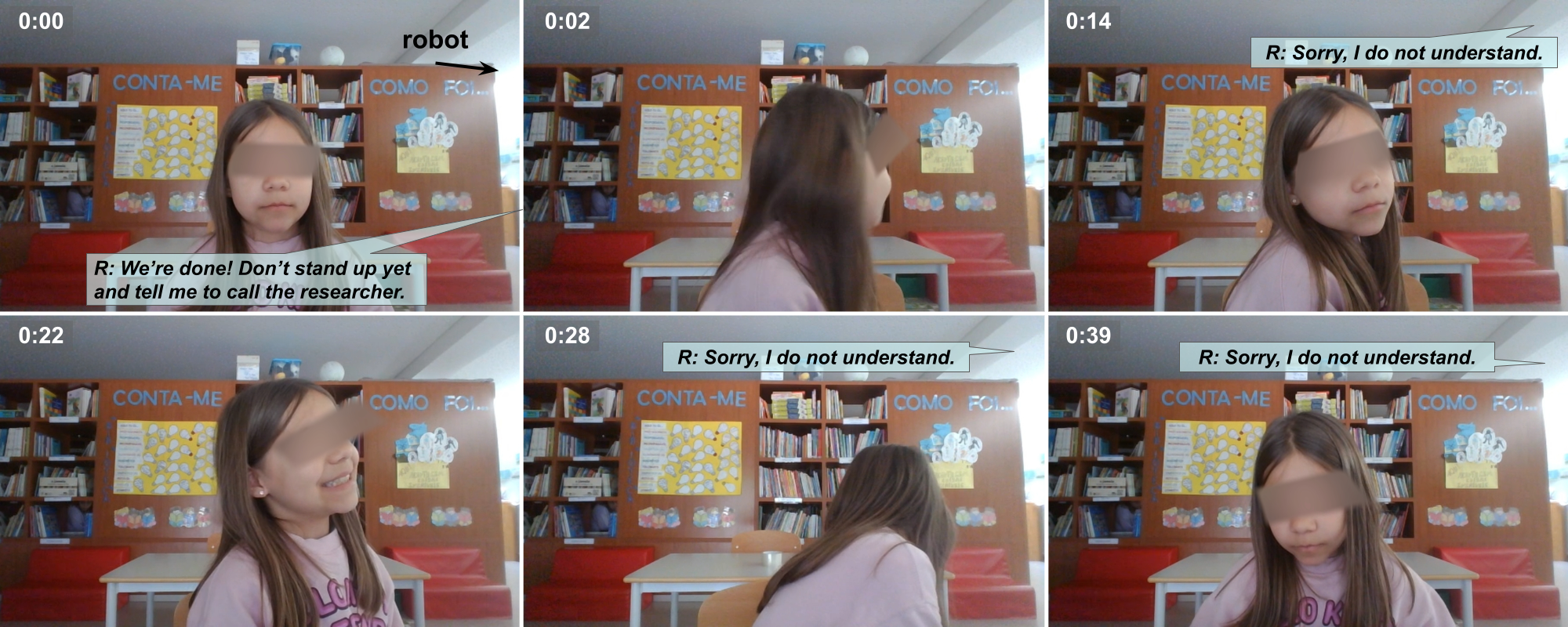}
    \caption{Absence of verbal engagement. In this vignette, the child ignores the robot's instructions and immediately \textcolor{disengagement}{looks for the researcher}. Through repeated error, child exhibits \textcolor{emotion}{confusion} (0:14), \textcolor{emotion}{amusement} (0:22) and continues \textcolor{disengagement}{looking back} or \textcolor{disengagement}{looking at the screen}.}
    \Description{Absence of verbal engagement. In this vignette, the child ignores the robot's instructions and immediately looks for the researcher. Through repeated error, child exhibits confusion (0:14), amusement (0:22) and continues looking back or looking at the screen.}
    \label{fig:vignettep14}
\end{figure*}

\paragraph{\textbf{Disengagement Behaviors}}
A key finding was children's progressive disengagement from robot interaction. Disengagement behaviors increased markedly after Error II, with ``no prompt'' responses and ``looking for researcher'' being most frequent (\autoref{fig:disengagement_frequency}). Some children completely discontinued interaction, leading to early experiment termination where not all three errors occurred (\autoref{fig:vignettep14}).

\begin{figure}
    \centering
    \includegraphics[width=0.9\linewidth]{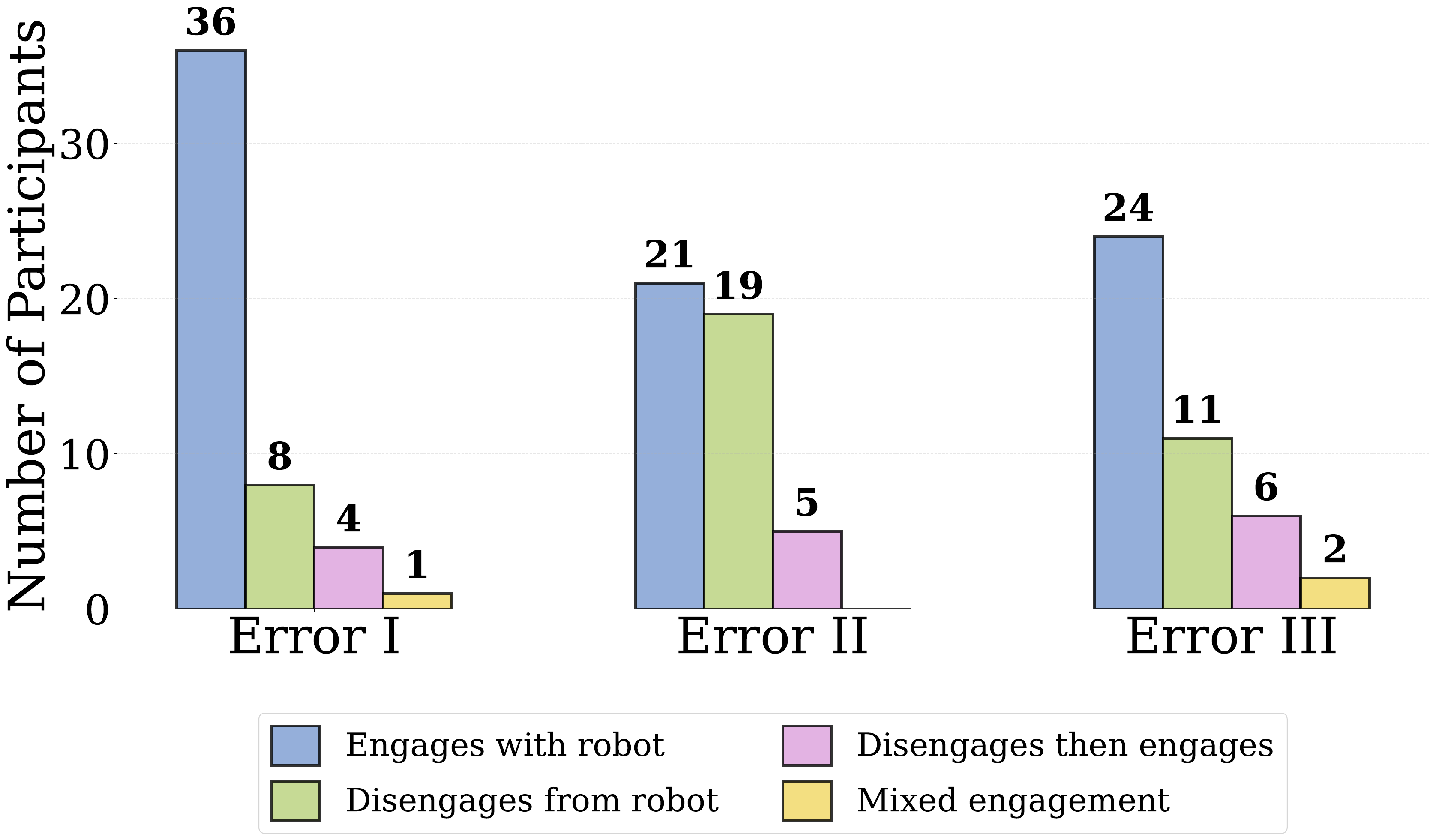}
    \caption{Children's engagement patterns across successive robot errors. Blue bars represent continued robot engagement, green bars show disengagement from robot interaction, pink and yellow show complex engagement patterns: that is, for a single error, children might initially exhibit disengagement (e.g., looking for the researcher) and then engage with the robot (e.g., by speaking to it); \textit{mixed} engagement means children switch behavior types more than two times. Notable shift from predominantly engaging behavior in Error I to mixed patterns in Errors II and III.}
    \Description{Children's engagement patterns across successive robot errors. Blue bars represent continued robot engagement, green bars show disengagement from robot interaction, pink and yellow show complex engagement patterns: that is, for a single error, children might initially exhibit disengagement (e.g., looking for the researcher) and then engage with the robot (e.g., by speaking to it); mixed engagement means children switch behavior types more than two times. Notable shift from predominantly engaging behavior in Error I to mixed patterns in Errors II and III.}
    \label{fig:engagement_breakdown}
\end{figure}

\paragraph{\textbf{Engagement Pattern Dynamics}}
To understand children's engagement dynamics, we investigated the engagement-disengagement patterns across each successive error. We categorize \textit{engagement} as any verbal interaction with the robot (including tone/cadence changes or moving closer), while \textit{disengagement} includes behaviors where attention shifted away from the robot (standing up, seeking researcher, leaving room). Most children initially engaged but showed varied patterns across errors (\autoref{fig:engagement_breakdown}). The three most common engagement trajectories (behaviors for Error I, Error II, and Error III, respectively) were:
\begin{itemize}
    \item Engage → Engage → Engage ($n=15$)
    \item Engage → Disengage → Engage ($n=5$)  
    \item Engage → Disengage → Disengage ($n=4$)
\end{itemize}

\paragraph{\textbf{Response Timing Changes}}
Finally, we investigated response timing -- that is, how long it took, after each robot error (``sorry, I do not understand''), for the child to exhibit a response.\autoref{fig:response_times} shows the results of our analysis. Response latencies increased across successive errors, with Error III showing the longest and most variable response times, suggesting children required more processing time as failures accumulated.

\subsubsection{Successive Performance Error in Children and Adults}

\paragraph{\textbf{Similarities with Adult Behavior}}

Several response patterns aligned with previous adult findings~\cite{liu2025successive}:
\begin{itemize}
    \item \textbf{Verbal adaptation}: Both children and adults modified prompts and adjusted vocal tone/cadence when encountering failures.
    \item \textbf{Emotional progression}: Similar evolution from confusion to frustration across successive errors.
    \item \textbf{Response timing}: Both populations showed increased response latencies and variability with successive failures.
    \item \textbf{Interaction abandonment}: Some participants in both studies completely discontinued interaction by the third error.
\end{itemize}

\paragraph{\textbf{Key Differences from Adult Behavior}}
Children exhibited several distinct behavioral patterns:
\begin{itemize}
    \item \textbf{Greater disengagement}: Children showed notably more disengagement behaviors, often temporarily ignoring the robot.
    \item \textbf{External help-seeking}: Unlike adults, children looked for or called the researcher, demonstrating different agency expectations in problematic interactions.
    \item \textbf{Politeness strategies}: Children more readily shifted to polite language forms (``please'') when initial commands failed.
    \item \textbf{Earlier disengagement}: While adults typically abandoned interaction primarily after Error III (n=7), children showed disengagement beginning at Error I.
    \item \textbf{Mixed engagement patterns}: Children displayed more dynamic engagement patterns, sometimes disengaging then re-engaging, unlike the more linear progression observed in adults.
\end{itemize}

These differences suggest that children approach robot failure recovery with distinct expectations about authority, help-seeking, and social interaction norms compared to adult users.

%% file: sections/05_discussion.tex
\section{Discussion} \label{sec:discussion}

Our study's unique contribution lies in its systematic reproduction of the Liu et al. successive error paradigm with children, enabling direct adult-child behavioral comparisons under controlled conditions. While existing work examines error responses in naturalistic, long-term settings, our brief, controlled protocol isolates immediate reactions to successive failures and reveals distinct child-specific behaviors that have important implications for error recovery design. This complements longitudinal studies by providing a focused snapshot of early-stage responses, bridging the gap between adult and child HRI research. The design implications derived from our observations are summarized in \autoref{tab:design_implications_cri}.

\paragraph{\textbf{Robot Errors Did Not Change Perception of Robot}} 
Our questionnaire findings suggest that children's overall perception of the robot remained stable despite experiencing both social errors (interruption) and performance failures (repeated misunderstanding). This robustness aligns with prior research indicating that children's conversational expectations with robots are more flexible \cite{rudenko2024child}, and that interruption behaviors went largely unnoticed or ignored by our participants. This finding is also consistent with broader evidence that robot errors do not necessarily diminish children's robot perceptions \cite{UNLUTABAK2025information,stower2022robot}. \citet{stower2024wrong} found that children's perception of robot reliability operates separately from their perception of the robot as a social interaction partner, while \citet{lighart2022learning}found that only  20\% of children were reportedly ``annoyed'' with robot errors, while the remaining accepted these errors as the robot was ``still learning''. These studies hint that children may possess a developmental advantage, compared to adults, in maintaining positive robot relationships despite technical limitations. However, notably \citet{lighart2022learning} also report a negative correlation between recognition errors and comfort/social attractiveness, indicating potential longer-context relational effects.

\paragraph{\textbf{Children's Unique Response Patterns to Robot Failures}}
Children exhibited distinct behavioral patterns when encountering successive robot failures, combining universal human responses with developmentally specific strategies. While the progression from confusion to frustration mirrors adult reactions \cite{liu2025successive}, children demonstrated unique coping mechanisms. Their various reprompting strategies -- repeating, simplifying, or altering vocal tone -- are not merely a reaction to a bug; they are a display of conversational repair \cite{Turkle2006artifacts}. These behaviors demonstrate children actively applying learned social schemas to robot interactions, attempting to restore conversational coherence even when the robot appears conversationally incompetent. The adoption of polite language forms (``please'') when initial commands failed suggests children transfer human social politeness norms to robot encounters, reflecting their developing understanding of social hierarchies and appropriate communication strategies \cite{Garvey1981turntaking}.

\begin{table*}[t]
\centering
\small
\caption{Design implications for and around robot error in CRI, derived from this study and prior research \cite{liu2025successive, serholt2018longitudinal,stower2022robot,lighart2022learning,Veivo02025breakdowns}.}
\label{tab:design_implications_cri}
\begin{tabular}{|p{5cm}|p{12cm}|}
\hline
\textbf{Observed Child Reaction/Insight} & \textbf{Direct Design Implication for CRI} \\
\hline
Children’s overall robot perception remained stable despite successive errors. & \textbf{Design for Error Tolerance (Perceptual Resilience):} Focus on managing errors gracefully rather than aggressively eliminating them. Allow minor, non-critical errors to occur without immediate, elaborate apology, leveraging children's flexible expectations. \\
\hline
Children frequently look for or call the researcher/adult after repeated failures (Calling for Backup). & \textbf{Facilitate External Help-Seeking (Agency Transfer):} Recognize explicit requests for human assistance (verbal or non-verbal) as a valid adaptive strategy and implement graceful mechanisms for transferring agency or summoning human aid. \\
\hline
Children use sophisticated verbal strategies: repeating, increasing specificity, eventually resorting to polite requests ("please"). & \textbf{Implement Multi-Layered Conversational Repair:} Design auditory input systems that are highly sensitive to prompt reformulation (changes in cadence, tone, linguistic complexity) and increase error confidence based on politeness markers (e.g., "please"). \\
\hline
Children exhibit dynamic engagement patterns, cycling between disengagement and re-engagement with the robot. & \textbf{Employ Non-Intrusive Error Recovery (Tolerate Pauses):} Recognize disengagement as an active problem-solving phase. Avoid immediate, intrusive re-engagement protocols; instead, monitor passively and allow the child to naturally re-engage.  \\
\hline
Response latencies increase with successive errors. & \textbf{Utilize Timing as Failure Metric:} Integrate response latency analysis (time elapsed between robot error and child's subsequent response) as a real-time, objective measure of escalating cognitive load and distress.  \\
\hline
\end{tabular}
\end{table*}

\paragraph{\textbf{Calling for Backup}}
A critical distinguishing feature was children's frequent external help-seeking behavior, particularly looking for or calling the researcher. This ``calling for backup'' pattern is consistent with previous reporting \cite{Veivo02025breakdowns} in a robot-assisted language learning scenario, and diverges markedly from adult participants who predominantly attempted self-directed problem-solving \cite{liu2025successive}, and likely reflects developmental variations in perceived agency and authority relationships, where children expect adults to provide solutions to complex problems. \citet{serholt2020repair} similarly documented children's responses to robot errors, including moving closer to robots, verbal adaptation, and importantly, disengagement and help-seeking. Rather than representing interaction failure, these behaviors demonstrate children's adaptive problem-solving and have important implications for educational or therapeutic contexts, where help-seeking could be leveraged as positive engagement indicators.

\paragraph{\textbf{Progressive Disengagement and Re-engagement Dynamics}}
Children's engagement patterns proved more dynamic than the linear adult progression, with participants cycling between engagement and disengagement states. While \citet{oertel2020engagement} report that children may disengage from robot interactions when novelty effects diminish, our short interaction duration and observed re-engagement patterns suggest different mechanisms. Instead, we interpret these disengagement patterns as active problem-solving and adaptation. When a tool fails to work as expected, a competent user may abandon it and seek an alternative strategy to accomplish their goal. In this case, the children's primary goal was to get the researcher to re-enter the room. The robot was merely the tool to achieve this goal. When the robot failed repeatedly, the children rationally disengaged from the faulty tool to pursue a more effective strategy: seeking a human for help. This reframes the behavior from ``robot failure'' to ``user competence,'' demonstrating sophisticated meta-cognition where children assess tool limitations and adapt their problem-solving approaches. Adults, conversely, may have been inhibited from displaying these do-it-yourself behaviors due to a more developed understanding of the context of the interaction as a scientific study. The engagement-disengagement flexibility suggests children are resilient to robot failures, maintaining openness to interaction even after experiencing repeated errors \cite{serholt2018longitudinal}.



\paragraph{\textbf{Methodological Limitations and Contextual Constraints}}
The highly controlled interaction setting with a specific conversational failure paradigm represents only one type of robot error scenario; our findings, while supported by other literature, are highly contextual to the experimental setting and children's developmental stage. Additionally, data collection in public schools required substantial experimental adaptation and reduced environmental control, including unexpected interruptions that led to participant exclusions 
Children's variable compliance with experimental instructions, with some choosing to ignore robot interaction bids entirely, reflects the authentic challenges of pediatric research but complicates systematic analysis. The limited age range and cultural homogeneity of participants further constrain transferability \cite{rudenko2024child,spitale2023systematicreviewreproducibilitychildrobot}. Most critically, the brief interaction duration during successive errors (average 36.28 seconds) provides only a snapshot of child-robot dynamics, and extended interactions might yield substantially different adaptation patterns. 


\paragraph{\textbf{Ethical Considerations and Data Protection Complexities}}
The study highlighted substantial challenges in international collaborative research involving children's data, particularly when data collection occurs under European data protection frameworks. Stringent requirements for children's data protection, while absolutely necessary for participant welfare, created significant barriers to data sharing and collaborative analysis. These protections ensure appropriate safeguarding of vulnerable participants but simultaneously limit opportunities for cross-cultural validation and broader collaborative insights. 

This study also employed mild deception through wizard-of-oz control, where children believed the robot was autonomous while researchers controlled its failures. While such deception is sometimes necessary to elicit naturalistic responses and is ethically acceptable when properly managed, it requires careful consideration with child participants. All children received thorough debriefing explaining the robot's true operation and the study's purpose, ensuring they understood no actual malfunction occurred and that their reactions were valuable contributions to improving robot design. Future research involving deception with children should continue to prioritize transparent debriefing procedures and ensure that the temporary deception does not cause lasting confusion or distress about technology capabilities.

\paragraph{\textbf{Future Directions and Broader Applications}}
Future research should explore longitudinal development of children's error tolerance and recovery strategies, investigate cultural variations in robot error responses, and examine different error types (social and performance) to develop more inclusive design practices. The observed help-seeking behaviors warrant investigation as potential positive indicators for human-robot collaborative learning environments.

A significant practical challenge arose when attempting to automate error detection through children's behavioral signals. Unlike adults who remain seated and within the camera frame, children frequently moved unpredictably, standing, turning, or positioning themselves at unexpected angles. This dynamic behavior created substantial missing data for computer vision and pose estimation algorithms, rendering traditional facial expression and body pose analysis unreliable. Future automated systems for child-robot interaction will require specialized approaches, including multiple camera angles, robust tracking algorithms handling occlusion and rapid movement, and potentially alternative sensing modalities beyond visual analysis. These constraints highlight the need for child-specific behavioral signal processing that accounts for children's inherently more dynamic and unpredictable physical behavior.


%% file: sections/06_conclusion.tex
\section{Conclusion} \label{sec:conclusion}

This study provides an examination of how children respond to successive robot failures. The observed help-seeking behaviors and perceptual resilience suggest children may be more adaptable partners in human-robot collaboration than previously recognized, though this comes with unique technical challenges for automated error detection systems. As robots increasingly enter social contexts involving children, these insights become crucial for creating effective, sustainable, and developmentally appropriate human-robot interaction systems.